\documentclass[runningheads]{llncs}
\usepackage{graphicx}
\usepackage{makeidx}
\usepackage{booktabs}
\usepackage{tabularx}
\usepackage{pifont}
\usepackage{amsmath, amssymb}
\usepackage{multirow}

\newcommand{\cmark}{\ding{51}}%
\newcommand{\xmark}{\ding{55}}%

\begin{document}
\title{Contrastive Learning for View Classification of Echocardiograms}

\author{Agisilaos Chartsias\inst{1} \and
Shan Gao\inst{1} \and
Angela Mumith\inst{1} \and 
Jorge Oliveira\inst{1} \and 
Kanwal Bhatia\inst{1, 2} \and
Bernhard Kainz\inst{1, 3} \and
Arian Beqiri\inst{1, 4}
}
\authorrunning{A. Chartsias et al.}

\institute{Ultromics Ltd., 4630 Kingsgate, Cascade Way, Oxford Business Park South, Oxford, OX4 2SU, UK \and
Metalynx Ltd., 71-75 Shelton Street, London, WC2H 9JQ, UK \and
Imperial College London, 180 Queen's Gate, London, SW7 2AZ, UK
\and
King's College London, School of Biomedical Engineering \& Imaging Sciences, London, SE1 7EU, UK\\
\email{
agis.chartsias@ultromics.com}
}
\maketitle             

\begin{abstract}

Analysis of cardiac ultrasound images is commonly perfor\-med in routine clinical practice for quantification of cardiac function. Its increasing automation frequently employs deep learning networks that are trained to predict disease or detect image features. However, such models are extremely data-hungry and training requires labelling of many thousands of images by experienced clinicians. Here we propose the use of contrastive learning to mitigate the labelling bottleneck. We train view classification models for imbalanced cardiac ultrasound datasets and show improved performance for views/classes for which minimal labelled data is available. Compared to a na\"ive baseline model, we achieve an improvement in F1 score of up to 26\% in those views while maintaining state-of-the-art performance for the views with sufficiently many labelled training observations. 

\end{abstract}
\section{Introduction}

\begin{figure*}[t]
\centering
\includegraphics[width=1\textwidth]{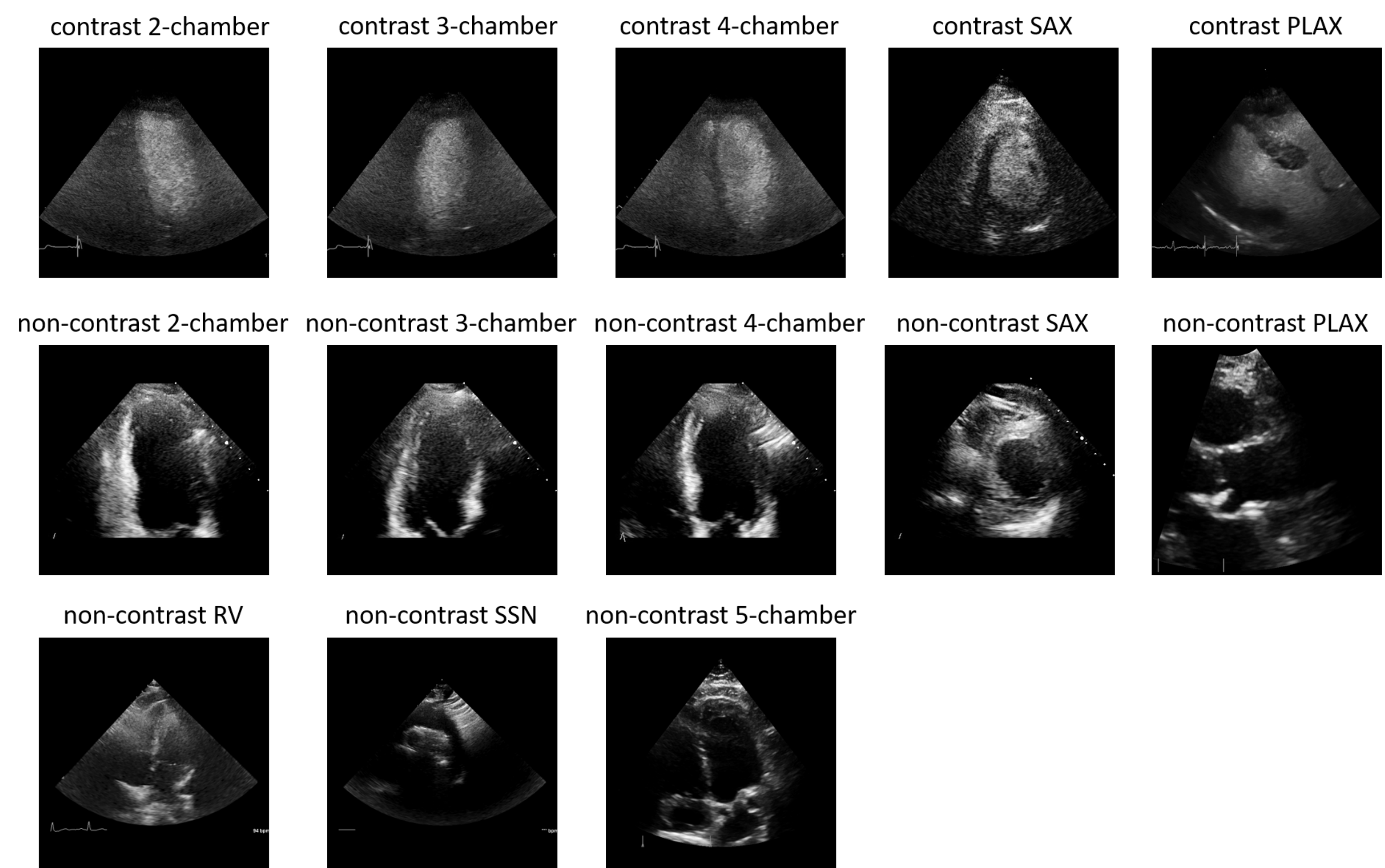}
\caption{Examples of different echocardiography views used including the 2/3/4/5 chamber apical, parasternal long-axis (PLAX), short-axis (SAX) at papillary muscle, right ventricular (RV) and suprasternal notch (SSN) views. The top row shows images obtained after injection of a microbubble contrast agent, causing a near inversion in image contrast, whereas the lower two rows show non-contrast images.}
\label{fig:ultrasound_example_views}
\end{figure*}

Echocardiography is widely and routinely used for assessing heart function and for the diagnosis of several conditions, such as heart failure and coronary artery disease~\cite{Lang2015}. In a routine echocardiographic study, multiple views of the heart are obtained to show different parts of the heart's internal structure, i.e. the ventricles, atria and valves --- see Figure~\ref{fig:ultrasound_example_views}. However, not all views are used in subsequent analysis of the echocardiograms depending on the cardiac function being assessed or the type of disease being investigated~\cite{Lang2015}. Therefore, an important initial step in any automated analysis pipeline is the accurate detection of standardised cardiac views shown on each echocardiogram. Frequently, further analysis --- usually performed with proprietary analysis software --- focuses on left ventricular function~\cite{Nagueh2016}. Often only the three apical views of the heart are assessed, which show slices through the left ventricle. 
However, it is still important for a view classifier to be aware of the entire cardiac anatomy so that it does not misclassify views it has not been trained on. This is challenging because it requires large training datasets with appropriate labels. 
Furthermore, when assessing certain cardiac conditions, the injection of a microbubble contrast agent is used to better highlight the boundaries of the left ventricular wall~\cite{PELLIKKA20201}. 
This changes the image appearance completely and effectively inverts the image contrast. Hence, these views cannot be classified without contrast enhanced data also being labelled for model training. 
The ability to correctly classify contrast images thus requires double the labelling effort.

View classification on echocardiographic data has previously been achieved using convolutional networks~\cite{gao2020fully,ostvik2019real,zhang2018fully} that take as input an image and predict one of the possible views that were present in the training label set for that network. For the commonly acquired echocardiographic views, such as the apical four-chamber view, labelled data for model training is available even in some public datasets~\cite{Leclerc2019,Ouyang2020}.
However, for less commonly acquired views, with or without contrast enhancement, it is time-consuming and expensive to acquire labels and thus, datasets are often highly imbalanced.  
To tackle data imbalance, training classifiers may require under-sampling the majority classes and specialised cost functions~\cite{johnson2019survey} or augmentations with synthetically generated data~\cite{antoniou2017data}.
 
In this paper, we investigate the problem of view classification in cardiac ultrasound images and attempt to improve the classification accuracy of convolutional neural networks, especially on under-represented classes, with the use of contrastive learning. Contrastive learning is a pre-training methodology, which improves learning of features useful for classification tasks through a contrastive loss. The contrastive loss clusters similar images together (positive pairs) and pushes different images away (negative pairs). This can be entirely based on self supervision for example when positive pairs consist of different augmented version of an image (SimCLR~\cite{chen2020simple}) or, when in addition to augmentations, positive pairs also use supervision to include images of the same label (SupCon~\cite{khosla2020supervised}). This has proven successful in computer vision tasks for instance for ImageNet sample classification~\cite{chen2020simple}.

Furthermore, although cardiac ultrasound data consist of videos, view classification is typically performed per-frame as a 2D classification problem. For videos, unsupervised contrastive learning, such as SimCLR, is not directly applicable as also discussed in~\cite{chen2020effective}: if multiple frames of the same video end up in the same batch, then the negative pairs of a frame will include other frames of the same video. This would hinder the ability of the contrastive loss to only cluster similar images together, since different video frames would generate a higher loss value.
We therefore adopt the supervised contrastive loss~\cite{khosla2020supervised}, which does not suffer from this limitation.
Our contributions are the following: (a) we apply contrastive classification neural networks to cardiac ultrasound, and (b) we evaluate in a dataset of contrast and non-contrast enhanced echocardiographic images collated from public and proprietary sources and show improved results when using the proposed contrastive framework for views which have fewer labelled  training observations.

\section{Related work}

Standard plane/view detection has been previously studied in fetal ultrasound with supervised deep learning models, such as SonoNet~\cite{baumgartner2017sononet}, multi-scale DenseNet \cite{kong2018automatic}, and convolutional networks finetuned with transfer learning~\cite{chen2015standard} or trained with additional tasks to predict attention maps and adversarial training~\cite{cai2018multi}. In echocardiography, inception~\cite{ostvik2019real} and VGG~\cite{zhang2018fully} networks have been used to predict several views or subclasses of views, although not applied on contrast echo data. Typically, contrast-enhanced images are used in isolation, for example to extract myocardial segmententations~\cite{li2017fully,li2021deep}. Most recently, high view classification accuracy was reported by a convolutional network applied on mixed microbubble contrast-enhanced and non-contrast data from a multi-vendor site~\cite{gao2020fully}.

Given sufficiently large datasets, supervised training of convolutional networks is successful in accurate view detection. 
However, network initialisation is important to facilitate convergence, and therefore pre-training methods using self-supervision with different augmented views of the same image~\cite{chen2020simple} or labels~\cite{khosla2020supervised} are investigated to improve computer vision classification tasks, such as on the ImageNet dataset.  
Contrastive learning has also been used in the medical domain, for instance to improve segmentation performance on MRI images~\cite{chaitanya2020contrastive} or to learn joint representations of ultrasound videos and speech~\cite{jiao2020self}.

\begin{figure*}[t]
\centering
\includegraphics[width=1\textwidth]{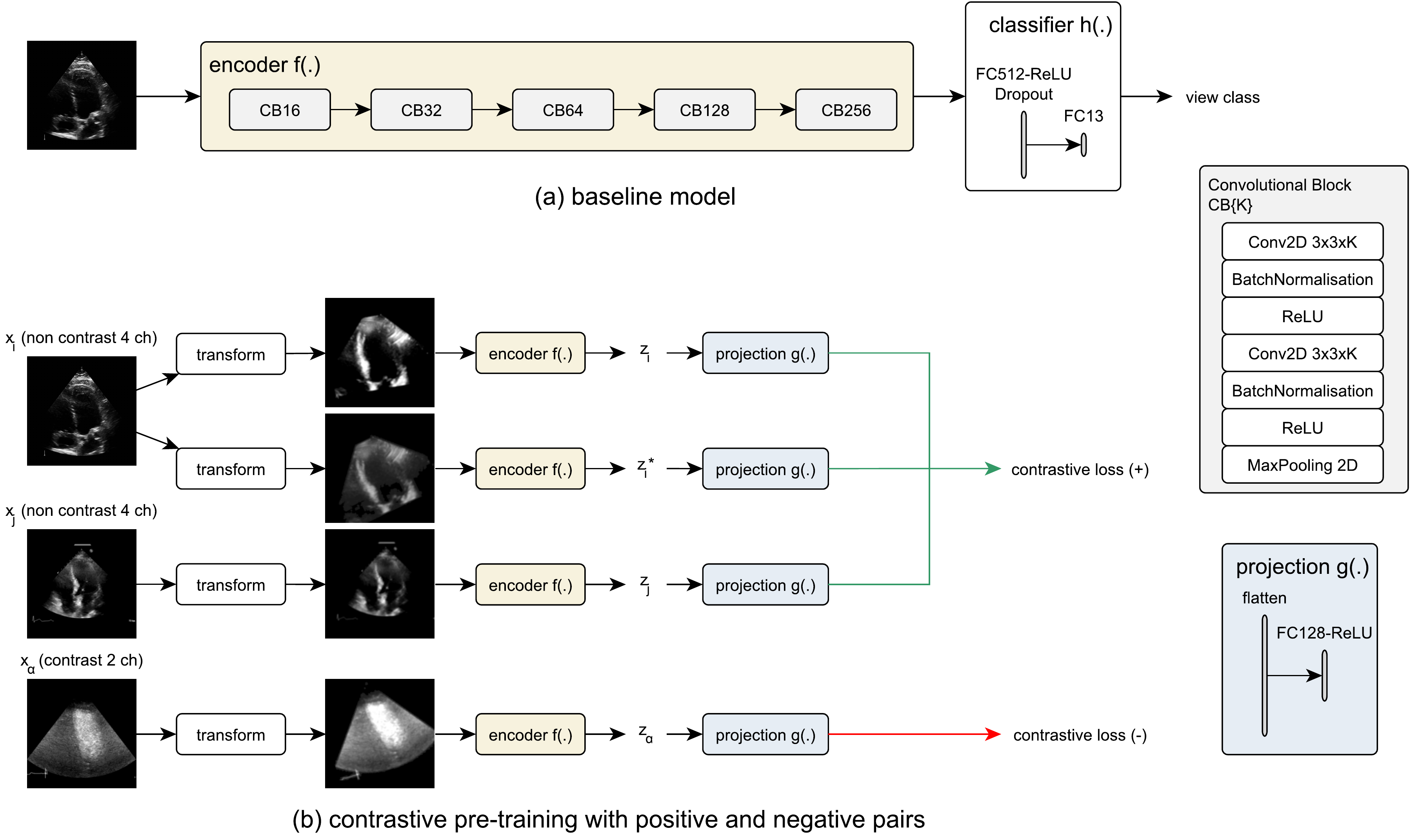}
\caption{Schematic of the baseline and contrastive models. (a) The baseline model architecture consists of a fully convolutional encoder and a fully connected classifier, and is trained with full supervision. (b) The contrastive model pre-trains the encoder using a projection network and a contrastive loss. The contrastive loss considers positive pairs if these are different augmentations of the same image or belong to the same class, and negative pairs otherwise.}
\label{fig:contrastive_schematic}
\end{figure*}

\section{Methodology}

Given an image $x$ of view $y_k$, where $k \in [1,13]$, corresponding to 13 classes of commonly acquired views with or without contrast, we consider a 2D baseline classification neural network $c(x)$ to detect per-frame view labels. This network maps input images through five convolutional blocks, each containing two convolutional layers followed by batch normalisation and a ReLU activation function, and a max pooling layer,
to a vector representation, which is then processed by two fully connected layers to generate a view label prediction. This model architecture, which was used in an eight-class form in \cite{gao2020fully}, is designed so that it is sufficiently small and effective on standard view classification and can be seen in Figure~\ref{fig:contrastive_schematic}a. Training is performed with the categorical cross entropy loss described as follows:
\[
L_{view}=-\sum_{k=1}^{13} y_k log(c(x)).
\]
A contrastive learning framework is then implemented as per the SupCon~\cite{khosla2020supervised} methodology as follows: we split the baseline model into a fully convolutional and a fully connected sub-model, which are used as an encoder $f(.)$ and classification network $h(.)$, respectively so that $c=h \circ f$. We add a projection network $g(.)$, which projects the encoded features $z=f(x)$ into a representation $\hat{x}=g(z)$. The projection $\hat{x}$ is used as an input to the contrastive loss that pre-trains the encoder.
Finally, the classification network $h(.)$
learns a mapping of the encoded features to their corresponding labels and is trained on a second stage following the encoder pre-training, whilst keeping the encoder weights fixed. A schematic of the framework is shown in Figure~\ref{fig:contrastive_schematic}.

The contrastive learning process is more formally described as: given $N$ randomly augmented images $\{x_i\}_{i=1}^N$, we first obtain a batch of $2N$ images $B=\{1\ldots2N\}$ by applying a second augmentation. For every image $x_i$ in the batch, and its projection $\hat{x_i}=g(f(x_i))$, there are also $M_i$ other images of the same label in the set $P_i=\{x_j\}_{j=1}^{M_i}$. According to~\cite{khosla2020supervised} the supervised contrastive loss is defined as:
\[
L_{supcon} = \sum_{i\in B}-log\left\{ \frac{1}{M_i}\sum_{j \in P_i} \frac{exp(\hat{x}_i \hat{x}_j / \tau)}{\sum_{\alpha \in B \backslash i} exp(\hat{x}_i \hat{x}_{\alpha} / \tau)} \right\},
\]
where $\tau$ controls the temperature scaling of the softmax. We set $\tau$=1000 as per~\cite{khosla2020supervised} and use brightness and contrast augmentations, as well as rotations to 30 degrees and spatial translations at up to 10\% of the image dimensions.

\begin{table}[b!]
    \centering
    \caption{Description of the training and test dataset.}
    \begin{tabular}{cc|ccc|ccc}
        \toprule
        \multirow{2}{*}{\textbf{Contrast}} & \multirow{2}{*}{\textbf{View}} & \multicolumn{3}{c|}{\textbf{Training set}} & \multicolumn{3}{c}{\textbf{Test set}} \\
        & & Subjects & Echocardiograms & Images & Subjects & Echocardiograms & Images \\
        \midrule
        \cmark & 2 ch.  & 711 & 1401 & 41603 & 139  & 276  & 5784  \\
        \cmark & 3 ch.  & 699 & 1377 & 41432 & 139  & 276  & 5763  \\
        \cmark & 4 ch.  & 704 & 1375 & 41214 & 138  & 274  & 5684  \\
        \cmark & plax   & 85  & 85   & 4547  & 9    & 9    & 560   \\
        \cmark & sax    & 607 & 1179 & 34387 & 138  & 275  & 5649  \\
        \xmark & 5 ch.  & 165 & 165  & 14714 & 18   & 18   & 1832  \\
        \xmark & plax   & 383 & 383  & 33969 & 42   & 42   & 3542  \\
        \xmark & rv     & 52  & 52   & 6521  & 5    & 5    & 703   \\
        \xmark & ssn    & 55  & 55   & 3483  & 6    & 6    & 336   \\
        \xmark & 2 ch.  & 314 & 544  & 26613 & 126  & 217  & 7061  \\
        \xmark & 3 ch.  & 364 & 605  & 32263 & 135  & 226  & 8662  \\
        \xmark & 4 ch.  & 332 & 556  & 28569 & 130  & 205  & 7938  \\
        \xmark & sax    & 229 & 437  & 17704 & 98   & 187  & 4234  \\
        \bottomrule
    \end{tabular}
    \label{table:dataset_description}
\end{table}

\subsection{Data} \label{sec:data}

The dataset used in this work comprised of anonymised 2D echocardiograms from multiple sites. The dataset is composed of data from EVAREST~\cite{Woodward2021}, a multi-site, multi-vendor UK trial,  some data from the EchoNet public dataset~\cite{Ouyang2020}, and some proprietary data from other imaging sites. The final dataset is split into a training and a test set of echocardiograms corresponding to 1,538 and 359 subjects, respectively. The total number of image frames contained in these data is 327,019 for the training set and 57,648 for the test set. Each echocardiographic video was labelled into one of 13 classes, which cover a set of standard cardiac views with or without microbubble contrast. 
The classes are shown in the first and second columns of Table~\ref{table:dataset_description} along with the number of subjects, echocardiograms and images present for each view.

Images were extracted from DICOM or AVI files and were pre-processed to remove all text information and annotations outside the ultrasound sector,  
so that the dataset contains only the images within the ultrasound sector.

As part of the EVAREST trial data, the dataset contains echocardiograms obtained with the patient at rest and with patients subjected to exercise or pharmacological stress. Heart rates vary from 45 to 150 and the number of heartbeats per scan are between one and three. The inclusion of stress echo data ensures that a range of image qualities is present in the dataset as stress echocardiograms tend to include images of poor image quality.

\section{Experiments and discussion} \label{sec:experiments}

\subsection{Experimental setup}

Prior to being fed into the network, image frames are resized to $192\times192$ pixel size, z-score normalised, and rescaled to $[0,1]$ range. The model and pipeline was developed in Python 3.7.7 with Tensorflow 2.2 and training was performed on four Nvidia GeForce RTX 2080 Ti GPUs with 11GB VRAM each.

The baseline and contrastive learning methods were trained using Adam with batch size 64\footnote[1]{The effective batch size is 128, since every image is augmented twice in a batch.} and learning rate equal to 0.0001 on a 8-fold cross-validation with the validation set containing 10\% of the training dataset's echocardiograms. Training stopped using an early stopping criterion based on the validation set.

We train models using all 13 view classes in two scenarios: one using all data, and then one with reduced data of around 50 echocardiograms per class, chosen at random. We report the mean F1 score, precision and recall across the different validation splits and a held out test set that is common across the different splits.

\begin{table}[t!]
\centering
\caption{Classification results (mean and standard deviation) of baseline and contrastive models on validation (taken from 10\% of the training set) and test sets using two datasets containing all data and 50 echocardiograms per class, respectively.}
\label{table:classification_results}
\begin{tabular}{l|lccc|ccc}
\toprule
\multirow{2}{*}{\textbf{Dataset}} & \multirow{2}{*}{\textbf{Method}} & \multicolumn{3}{c|}{\textbf{Validation set}} & \multicolumn{3}{c}{\textbf{Test set}} \\
 & & F1 Score & Precision & Recall & F1 Score & Precision & Recall \\
\midrule
50 echocardiograms  & {Baseline}      & $0.794_{.02}$ & $0.780_{.02}$ & $0.837_{.01}$ & $0.765_{.02}$ & $0.756_{.03}$ & $0.820_{.02}$ \\
per class    & {SupCon}   & $0.800_{.01}$ & $0.787_{.02}$ & $0.833_{.01}$ & $0.775_{.01}$ & $0.770_{.02}$ & $0.825_{.01}$ \\
\midrule
\multirow{2}{*}{all data} & {Baseline}      & $0.911_{.02}$ & $0.924_{.03}$ & $0.902_{.02}$ & $0.874_{.01}$ & $0.896_{.02}$ & $0.880_{.02}$ \\
                          & {SupCon}   & $0.915_{.02}$ & $0.928_{.01}$ & $0.908_{.02}$ & $0.892_{.01}$ & $0.907_{.01}$ & $0.896_{.01}$ \\
\bottomrule
\end{tabular}
\end{table}

\subsection{Classification performance} \label{sec:classification}

Table~\ref{table:classification_results} shows the mean and standard deviation of F1 score, precision and recall for the experiments on the full and reduced datasets. Both methods perform equally well on the dataset of 50 echocardiograms per class, which is balanced. We observe an improvement in test F1 score on the full dataset, which increases from 0.874 to 0.892, and smaller standard deviations in precision and recall.

Table~\ref{table:class_wise_result} reports the per-class test F1 score for the two datasets. When assessing the per-class classifier performance, it can be seen that the contrastive training has minimal effect for the model trained on 50 echocardiograms per class. When training on the full dataset, classes which have a larger number of training data show similar or marginal improvement in performance in the test set. However, classes with substantially less training data, such as the contrast PLAX view, non-contrast 5-chamber view, and the non-contrast right ventricular (RV) view show greater improvement when using contrastive learning. The non-contrast suprasternal notch (SSN) view shows a 4\% reduction but both baseline and contrastive model accuracies are very high.

\begin{table}[t]
\centering
\caption{Classification results (mean and standard deviation) per class. The first column indicates whether the images have contrast or not. Results show the F1 score on the test set for two experiments using different training set sizes, with the number of studies for each view shown. Highest differences are marked in bold.}
\label{table:class_wise_result}
\begin{tabular}{clcccc|cccc}
\toprule
\textbf{Cont} & \textbf{View} & \textbf{Size} & \textbf{Baseline} & \textbf{SupCon} & \textbf{\%Diff} & \textbf{Size} & \textbf{Baseline} & \textbf{SupCon} & \textbf{\%Diff} \\
\midrule
\cmark & 2 ch.     & 50 & $0.693_{.03}$ & $0.702_{.02}$ & 1.24  & 677          & $0.866_{.01}$ & $0.870_{.01}$ & 0.42  \\
\cmark & 3 ch.     & 50 & $0.811_{.02}$ & $0.811_{.03}$ & 0.02  & 664          & $0.966_{.00}$ & $0.968_{.00}$ & 0.22  \\
\cmark & 4 ch.     & 50 & $0.758_{.07}$ & $0.737_{.06}$ & -2.73 & 672          & $0.888_{.00}$ & $0.896_{.01}$ & 0.99  \\
\cmark & plax      & 50 & $0.608_{.16}$ & $0.733_{.05}$ & \textbf{20.61} & 68  & $0.570_{.08}$ & $0.719_{.09}$ & \textbf{26.08} \\
\cmark & sax       & 50 & $0.926_{.04}$ & $0.946_{.01}$ & 2.22  & 570          & $0.985_{.00}$ & $0.986_{.00}$ & 0.12   \\
\xmark & 5 ch.     & 50 & $0.546_{.05}$ & $0.542_{.04}$ & -0.76 & 132          & $0.660_{.05}$ & $0.706_{.05}$ & \textbf{6.98}  \\
\xmark & plax      & 50 & $0.952_{.03}$ & $0.959_{.02}$ & 0.83  & 306          & $0.972_{.01}$ & $0.974_{.01}$ & 0.15   \\
\xmark & rv        & 42 & $0.358_{.06}$ & $0.363_{.06}$ & 1.45  & 42           & $0.632_{.12}$ & $0.697_{.09}$ & \textbf{10.26}  \\
\xmark & ssn       & 44 & $0.700_{.06}$ & $0.679_{.04}$ & -3.03 & 44           & $0.990_{.03}$ & $0.952_{.04}$ & -3.88 \\
\xmark & 2 ch.     & 50 & $0.857_{.01}$ & $0.856_{.01}$ & -0.04 & 269          & $0.939_{.00}$ & $0.934_{.01}$ & -0.54  \\
\xmark & 3 ch.     & 50 & $0.912_{.01}$ & $0.910_{.01}$ & -0.24 & 319          & $0.967_{.01}$ & $0.969_{.00}$ & 0.22  \\
\xmark & 4 ch.     & 50 & $0.879_{.01}$ & $0.877_{.01}$ & -0.23 & 287          & $0.937_{.01}$ & $0.936_{.01}$ & -0.06  \\
\xmark & sax       & 50 & $0.951_{.01}$ & $0.963_{.01}$ & 1.27  & 213          & $0.988_{.00}$ & $0.989_{.00}$ & 0.10    \\
\bottomrule
\end{tabular}
\end{table}

\subsection{Ablation studies and failure cases}

We perform two ablation experiments on the model parameters. Firstly, we evaluate the effect of batch size  by testing values equal to 32 and 16. The obtained results are the same as the ones achieved with batch size 64. Although it has been reported that large batch sizes benefit contrastive learning~\cite{chen2020simple}, since more positive and negative examples occur in a batch, at this value range the effect is minimal. GPU memory limitations prevented experiments with higher values.

We also experiment with different sets of augmentations. The experiments in Section~\ref{sec:classification} use random rotations, translations, as well as changes in brightness and contrast. Random crops resulting in images of $140\times140$ pixel size have also been tested. However, training with such crop augmentations decreased the validation F1 score of the contrastive model by approximately 15\%. This can be attributed to the fact that in view classification, cropped ultrasound images might generate images which appear similar to other views.

\begin{figure*}[t]
\centering
\includegraphics[width=1\textwidth]{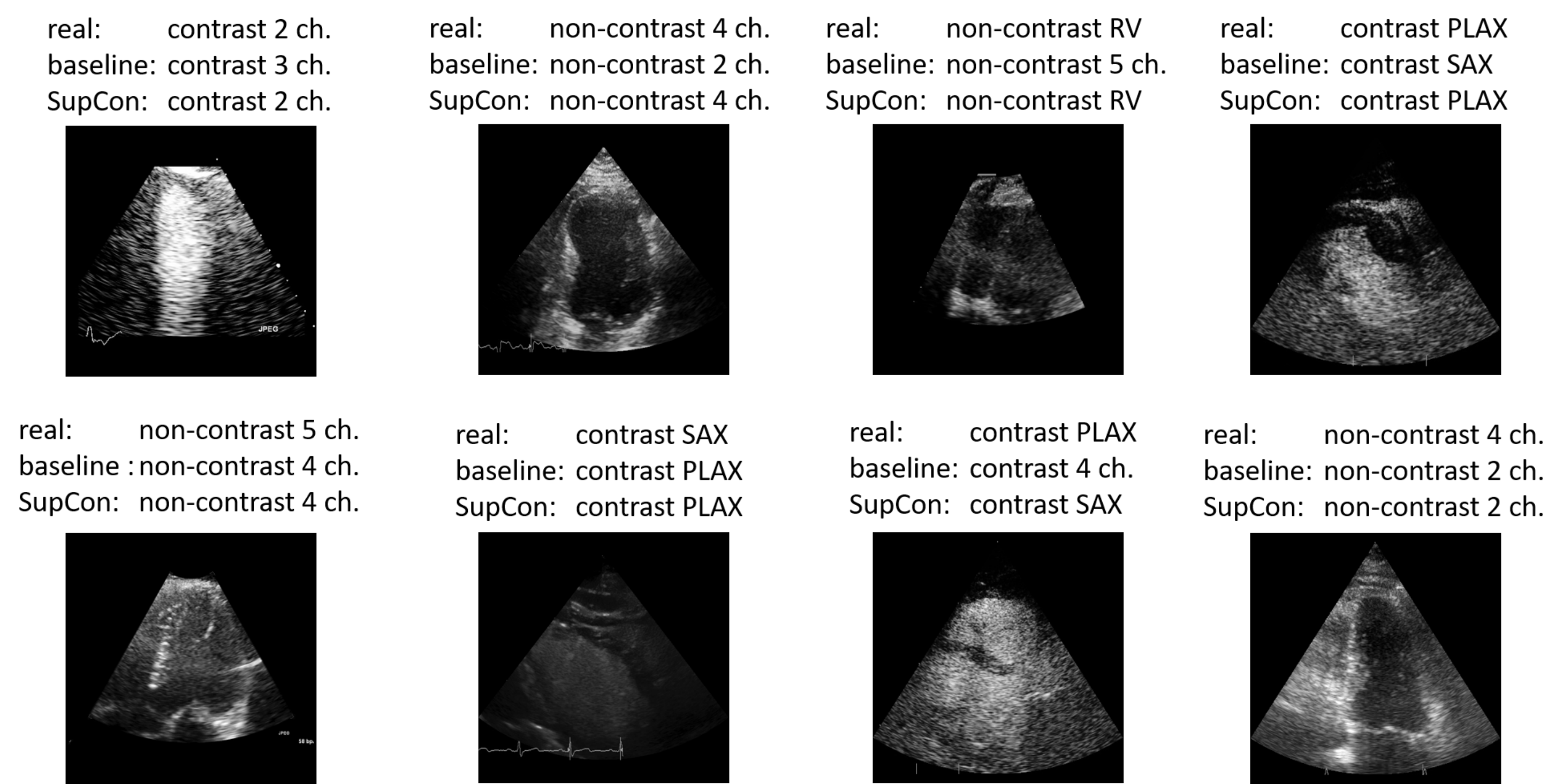}
\caption{Selection of failure cases. The baseline model fails on all these, but SupCon correctly classifies the examples in the top row.}
\label{fig:failure_cases}
\end{figure*}

Finally, Figure~\ref{fig:failure_cases} shows a selection of cases for which the baseline model fails, but for some the contrastive model is able to predict correctly. In all cases, the incorrect view is visually similar to the true view (for example, the apical 4 and 5 chamber views are very similar) so it is evident why the models would struggle. The contrastive model is likely more successful with these challenging views as it creates a better decision boundary between classes.

\section{Conclusion}
We have shown that the use of contrastive learning applied to echocardiographic view classification can improve accuracy and reduce standard deviation of the classifier for views for which far less training data is available, with no reduction in overall performance. This indicates that contrastive learning could be a powerful tool in developing models for analysing medical images without requiring such intensive collection and labelling of very large datasets. 

We leave as future work testing the effect of different contrastive losses on diverse datasets potentially including unlabelled data, as well as studying the effect of design biases introduced by different encoder architectures on the quality of the learnt latent representations.

\subsection*{Acknowledgements}

We thank the echocardiographers involved in this study for their thorough annotation of images from the EVAREST dataset.

\bibliographystyle{splncs04}
\bibliography{references}

\end{document}